# Real-Time Alert Correlation with Type Graphs

Gianni Tedesco[1] and Uwe Aickelin[1]

School of Computer Science,
University of Nottingham,
Nottingham NG8 1BB,
United Kingdom

**Abstract.** The premise of automated alert correlation is to accept that false alerts from a low level intrusion detection system are inevitable and use attack models to explain the output in an understandable way. Several algorithms exist for this purpose which use attack graphs to model the ways in which attacks can be combined. These algorithms can be classified in to two broad categories namely scenario-graph approaches, which create an attack model starting from a vulnerability assessment and type-graph approaches which rely on an abstract model of the relations between attack types. Some research in to improving the efficiency of type-graph correlation has been carried out but this research has ignored the hypothesizing of missing alerts. Our work is to present a novel type-graph algorithm which unifies correlation and hypothesizing in to a single operation. Our experimental results indicate that the approach is extremely efficient in the face of intensive alerts and produces compact output graphs comparable to other techniques.

## 1 Introduction

The output of intrusion detection systems (IDS) is generally a time series of discrete events called "alerts" with each event describing, at a low level, features of the network traffic. These alert attributes typically include the endpoints and communication channels implicated in an alert and the type of alert. Arguably the most significant problem with analyzing IDS alerts is the high volume of false alarms. Even without false alarms IDS alerts require some interpretation. This is because attacks are often split in to several stages, each of which may generate many alerts.

This observation has lead to the proposal that alerts be automatically correlated using a model of attacks which encodes their prerequisites and consequences [10]. Typically these methods involve representing attack types as vertices in a directed acyclic graph which we shall call an "attack graph". Edges in attack graphs represent the relationship between prerequisites and consequences of attacks. Intuitively speaking, a directed edge will connect attack A to attack B if A prepares for B.

Research has shown that such techniques are capable of:

1. Aggregating alerts which imply the same, or similar, consequences. An aggregated group of alerts is called a hyper-alert.

2. Ignoring extraneous alerts which do not correlate with anything.
3. Uncovering missing alerts in an alert stream and hypothesizing their attribute values where possible [8]. Hypotheses may optionally be compared against other evidence sources such as system logs [11].

These automated alert correlation techniques may be divided in to two categories based on the type of attack model which is encoded in the attack graph. We shall refer to the two categories as type-graph and scenario-graph algorithms. Scenario graph algorithms rely on a complete and correct vulnerability assessment to generate a graph of attack sequences specific to the protected network[6]. While this approach allows for real-time automated correlation it fails completely if network addresses are re-assigned or if the vulnerability assessment is erroneous. Conversely, type graph algorithms model only abstract attack types which allows for more robust correlation but with a higher computational cost. In [10] correlation is performed in batch mode only and in [15], where vulnerability assessment data is incorporated, a sliding correlation window is required to keep the problem manageable.

We assert that real-time correlation is desirable because it allows for timely automated responses. If the time lag between detection and response is too great then attacks such as rapidly spreading worms may become much more difficult to contain. Real-time operation also facilitates techniques such as [4] where correlation output is used to perform a targeted forensic analysis of network traffic for the purposes of discovering novel attacks and variations of known attacks.

Our work is motivated by the need for a correlation algorithm with both the flexibility of an abstract attack type-graph and similar performance characteristics to state of the art scenario graph algorithms. Specifically, we wish to avoid relying on prior knowledge of network topology and the distribution of vulnerabilities in the protected network. It is also desirable to avoid relying on a sliding correlation window which would allow "low and slow" attacks to become lost.

The aim of this paper is to develop an automated alert correlation algorithm using attack type graphs which is suitable for deployment in a real-time setting. A theoretical analysis of computational complexity will be provided. For verification the algorithm will be experimentally evaluated in terms of performance and accuracy.

Our proposed solution works by re-structuring the type graph correlation algorithm presented by Ning et al. such that it acts on individual alerts in sequence rather than all alerts in batch. The basic approach is to keep an internal database of hyper-alerts of each type and use in-memory indexes to efficiently find prerequisites of each new hyper-alert. The size of the in-memory database is minimized by eliminating redundant information which does not contribute to the correlation process. Hypothesizing of missing alerts is a recursive special-case of the correlation algorithm which can input hyper-alerts with wild-card attributes. The main contributions of this work are a type-graph correlation algorithm suitable for real-time use. The algorithm depends on a novel index structure and unifies the correlation and hypothesising steps in to a single algorithm.

This paper is structured as follows. First a brief discussion of related work is given in section 2. From here we present a description and formal definition of the problem in section 3. Building on this definition, a solution is presented in section 4 which solves the minimal IAC problem where there are no false negative alerts. This minimal algorithm is developed to the fuller solution presented and analyzed in section 5. Section 6 provides an empirical analysis of the algorithm. In the final section the results are discussed, conclusions drawn and future work proposed.

## 2 Related Work

Seminal works such as [5, 13] laid the groundwork for automated analysis of security related facts and events. These works proposed a formal theory of computer attacks by modeling the prerequisites and consequences of vulnerabilities in attack graphs and formal grammars respectively.

Wang et al. take a vulnerability-centric approach to alert correlation [6]. In this work an automated vulnerability analysis [12, 14] creates an attack graph consisting of two types of vertex, attacks and states. Only those attacks which have been found on the protected system are included. All attack vertices are bound with attribute values such as IP addresses and ports. The correlation algorithm works by performing a breadth first search on the attack graph. High performance is achieved by enumerating all possible fact assignments for every attack type and pre-computing an optimized graph structure for correlation. Another important concept in this work is "implicit correlation" whereby only the latest alert which satisfies an attack step is stored in memory. However, we have asserted that it is undesirable to assume that the defender can reliably know of all vulnerabilities on the network. Therefore our work uses an abstract attack-type model although we do use a similar hypothesizing technique and try to preserve the notion of implicit correlation as far as possible.

Ning et al. take a logical approach to modeling attack sequences for automated correlation[2, 9, 10]. The technique is intended to be applied in batch to an off-line database of collected hyper-alerts. The fundamental building block of the approach is the definition of a "hyper-alert type" which represents a type of attack and its prerequisites and consequences. Each hyper-alert type consists of a triple of fact names and prerequisites and conclusions which are predicate expressions with free variables bound from the fact names. If a predicate appears in the consequence of one hyper-alert type and the prerequisite of another then the former "may prepare for" the latter. The assignment of facts to any such shared predicate are used to calculate equality constraints between the two types.

A hyper-alert of a given type is simply a tuple of attribute values corresponding to the fact names for that type. Correlation is performed in batch on a set of hyper-alerts, each hyper-alert is considered a potential vertex in the correlation graph and if equality constraints are satisfied between other hyper-alerts then they are correlated by adding a directed edge between them provided that their timestamps show the correct temporal order.

Hypothesizing of missing alerts is treated in [8, 11]. The problem here is that when some steps in an attack have been missed by the underlying IDS then the resultant correlation graphs may be split and require additional processing to re-integrate them. The approach taken in their work involves four steps:

1. Subgraphs of the correlation graph are clustered according to the attribute values of their hyper alerts.
2. Once candidate subgraphs have been selected for integration, a special hyper-alert type-graph is consulted which has had indirect edges added to it. Pairs of subgraphs are then correlated using these new edges to define the indirectly-prepares-for relation.
3. When an indirect correlation occurs there are one or more paths in the type-graph connecting the two hyper-alert types. New hyper-alerts are created to connect the two correlation graphs and their attribute values inferred using the equality constraints in the graph.
4. Because the prior steps may generate many redundant hypotheses with equivalent fact values, a consolidation step reduces the size of the final correlation graph.

The work presented in this paper takes a different approach and simply relies on recursing backwards through the type-graph whenever a hyper-alert is input which has not had it's prerequisites met by another hyper-alert in the system. Our method is at once more efficient and eliminates the consolidation step by terminating recursion as soon as a duplicate hypothesis is generated.

In [15] correlation and hypothesizing is performed, again, in batch mode. However in this case a state/event model is chosen so that evidence from complementary sources such as vulnerability analysis and raw audit logs. The attack model is converted in to a Bayesian network where prior probabilities are assigned manually by human experts. A sliding time window is used to limit memory usage and prevent a combinatorial explosion in run-time complexity associated with the Bayesian inference algorithm.

Our work is most similar to [7] in which in-memory indexes are used to significantly speed up correlation leaving the RDBMS just to store a log of hyper-alerts on disk. The most relevant contribution in their work is the proposal to index instances of predicates rather than hyper alerts. Their results indicate that the algorithm would be suitable for real-time operation but hypothesizing of missing alerts is not addressed and must presumably be performed as a post-processing step on the correlation graph. The work presented in this paper takes a different approach and instead indexes instances of the $PrepareF or$ relation.

In summary, there are several automated correlation algorithms. Those which are suitable for real-time operation either rely on the defender being able to correctly and completely enumerate possible combinations of attacks on their protected network, or worse, rely on a sliding time window which opens up the correlator to "low and slow" or "alert injection" attacks. The abstract type-graph approach appears more promising and has been partly optimized for real-time deployment. Our work builds on prior techniques by using a novel indexing structure and unifying the correlation and hypothesizing steps in to a single real-time algorithm.

## 3 Problem Definition

For the purposes of clarity the intrusion alert correlation (IAC) problem will be solved in two steps. Firstly the "minimal IAC problem" in which a totally accurate alert stream is input and no alerts are hypothesized and secondly; the "extended IAC problem" in which some alerts can be missing and the system must hypothesize alerts. The following problem definition is based on that proposed by Ning et al.[8–11].

Definition 1. An attack model consists of logical predicates, hyper-alert types and implication relations. A hyper-alert type T is a triple (fact, prerequisite, consequence) where fact is a set of attribute names associated with the type, prerequisite and consequence are sets of predicate expressions with free variables bound from fact. $Prereq(T)$ and $Conseq(T)$ denote the set of predicate expressions from the prerequisite and consequence elements of T respectively. For brevity we assume all implied expressions to be included in $Conseq(T)$. We shall refer to the set of all hyper-alert types in an attack model as $\tau$.

For the purpose of our examples we will assume that there are always 4 elements in fact (say, source address, source port, destination address, destination port).

Definition 2. Given an ordered pair of hyper-alert types (A, B) then A may prepare for B if $Conseq(A)$ and $Prereq(B)$ share at least one predicate, with possibly different arguments.

Definition 3. Given an ordered pair of hyper-alert types (A, B) where A may prepare for B a set of equality constraints may be computed. Each such constraint is a set of logical conjunctions of equality comparisons between the attributes of the two types.

Let the sequences $u_1, u_2, ..., u_n$ and $v_1, v_2, ..., v_n$ be distinct facts in type A and B respectively. Then each constraint takes the form:

$$u_1 = v_1 \wedge u_2 = v_2 \wedge ... \wedge u_n = v_n$$

such that there exists $p(u_1, u_2, ...u_n) \in Conseq(A)$ and $p(v_1, v_2, ..., v_n) \in Prereq(B)$ where p is the same predicate with possibly different fact assignments.

Note that the only substantial difference between our definition and that of Ning et al. is the restriction that any given fact may appear at most once in the arguments of a predicate. The purpose of this restriction will become clear in the following sections.

**Definition 4.** Given an attack model, let us define an **attack-type graph** $TG = (V, E, C, T)$. Where $(V, E)$ is a directed acyclic graph. $T$ is a bijection of vertices on to hyper-alert types. An edge $e(v_0, v_1) \in E$ if and only if $T(v_0)$ may prepare for $T(v_1)$. $C$ maps each edge to a set of constraints.

**Definition 5.** A **hyper-alert** $h$ is simply a tuple of attribute values. $Type(h)$ is a mapping on to the set of hyper-alert types. $Prereq(h)$ and $Conseq(h)$ denote the set of predicates from the prerequisite and consequence of the hyper-alert type with free variables bound from the attribute values of the hyper-alert. $Timestamp(h)$ denotes the timestamp of the hyper-alert. A **hyper-alert stream** is any time-ordered series of hyper-alerts.

**Definition 6.** A hyper-alert $h$ of type A is said to **prepare for** a hyper-alert $h'$ of type B if and only if $Type(h)$ may prepare for $Type(h')$ and at least one equality constraint evaluates to true when fact names have been substituted with actual values from the hyper alerts. Furthermore, since an event B can be the cause of an event C if and only if B occurs before C, an implicit time constraint ensures forward causality holds. In other words the directed edges in TG, like time, move from past to future.

Two hyper alerts are said to be correlated if and only if the former prepares for the latter. Since all that is required to correlate two hyper alerts is that any one of the constraints holds. We might say that each edge in TG is labeled with a predicate logical formula, consisting only of equality comparisons, in disjunctive normal form.

**Definition 7.** The output **correlation graph** CG is $(V, E, H)$ where $(V, E)$ is a DAG and $H$ is a bijection of hyper-alerts to vertices and an edge $e(v_0, v_1) \in E$ if and only if $H(v_0)$ prepares for $H(v_1)$.

**Definition 8.** If a hyper-alert $h$ exists where $Prereq(h)$ is non-empty and there does not exist a hyper-alert $h'$ such that $h'$ prepares for $h$ then $h$ is said to be "unexplained".

An **unexplained alert** $h$ may sometimes be explained by the construction of a sequence of hypothesized hyper alerts $y_1, y_2, ..., y_n$ such that $y_n$ prepares for $h$, $y_{n-1}$ prepares for $y_n$, ..., and a real (unhypothesised) hyper alert $h'$ prepares for $y_1$. There may be several alternative explanations for any such hyper-alert.

The **extended correlation graph** EG therefore consists of $(V, E, H, Y)$ with the same definition as CG with the addition of $Y$, a mapping of vertices on to the set of hypothesised hyper-alerts which are required to explain any unexplained alerts in $H$. $V$ is formed by the union of $H$ and $Y$.

In summary our problem is to propose an algorithm which:

1. Is initialized with TG, and an empty CG.
2. At each time step:
   (a) Input a hyper-alert.
   (b) Construct a correct and complete CG as per definition 7 or, for the extended IAC problem, definition 8.

## 4  A Minimal Solution

The inner loop of our proposed algorithm consists of two steps. Firstly "searching for correlations" and secondly "marking of consequences". When marking consequences of a type T hyper-alert h we find each type T′ such that T may prepare for T′. Then the equality constraints between the two types are used so as to index every possible combination of hyper-alert attributes for T′ which should be considered prepared for by h. Each index entry created in this stage contains a pointer to h. Conversely when searching for correlations the indexes on type T are searched using the attributes of h. If an earlier hyper-alert h′ has been input and marked it's consequences it will be found during the searching for correlations stage if and only if h′ prepares for h. The structure of our index is unique and, by indexing each attribute combination separately, the IAC is reduced to a sufficiently small constant number of search and insert operations on balanced binary trees[1] rather than multi-dimensional searches with wild-cards.

This approach exploits two properties of the structure of the problem. Firstly that time flows from past to future, meaning that prior alerts do not need to be checked and correlated twice. Secondly although the number of possible constraints on a given edge are exponentially related to the number of facts, in practice, the number of facts and therefore the maximum number of indexes required is small.

**Lemma 1.** Given a pair of hyper-alert types $(T_0, T_1)$ we take A and B to be their attribute sets. The sets of attributes are equipotent, each containing n elements. Each constraint may be represented as a set containing $0 <= m <= n$ ordered pairs of attributes (a, b) such that $a \in A$ and $b \in B$. No element of A may appear as a left component more than once, and no element of B may appear as a right component more than once since by definition 3 the problem is restricted to the simplified case in which each fact referred to in an equality constraint may make at most one appearance on each side of the equation.

There are $P(n, m) \cdot C(n, m)$ ways to arrange m distinct pairs from n elements of A and n elements of B, where P and C are the permute and chose functions respectively. The number of possible equality constraints is therefore the sum of all constraints of each length m.

*Proof.* Our problem is to construct two sequences $a_1, a_2, ..., a_m$ and $b_1, b_2, ..., b_m$ where $a_1$ is paired with $b_1$, $a_2$ is paired with $b_2$, etc. We shall solve the problem in two separate steps. First we chose m elements of A and m elements of B and secondly we arrange the pairs. There are $C(n, m)^2$ ways to select a pair $(A′, B′)$ where $A′ \in$ the set of all m-combinations of elements in A and $B′ \in$ the set of all m-combinations of elements in B. Now to pair them up we keep elements of $A′$ in a fixed order and simply count the ways to permute the elements of $B′$. Since there are m! ways to permute m attributes:

$$C(n,m)^2 \cdot m! = \frac{n!}{m!(n-m)!} \cdot \frac{n!}{m!(n-m)!} \cdot m!$$
$$= \frac{n!}{m!(n-m)!} \cdot \frac{n!}{(n-m)!}$$
$$= C(n,m) \cdot P(n,m)$$

□

If we wish to count the maximum number of constraints when there is more than one type of attribute then we can re-use the formula above to count the ways of comparing the attributes of each type and take the product:

$$\prod_{i=1}^{t} \sum_{j=0}^{c_i} P(c_i, j) \cdot C(c_i, j) \quad (1)$$

Where t is the number of types, $c_i$ is the number of attributes of the $i^{th}$ type. Therefore, if we chose 4 facts: source and destination addresses and ports where addresses and ports are not comparable with each other. Then there are 49 possible constraints to an edge in TG. Since there are less combinations than permutations, the idea is to create an index for each of the 16 combinations of facts for each type. Permutations capture the possibly different orderings for the attributes in the equality constraints and will be used when inserting items in to the indexes.

Algorithm 1, requires several further definitions to determine which combinations of fields must be indexed for each type and how to evaluate what are the consequences for each hyper-alert so that they can be marked. A notion similar to implicit correlation in [6] is introduced. If two hyper-alerts have identical attribute values then they must also have identical consequences meaning that the correlation procedure is redundant the second time around. We define an implict correlation so that all hyper-alerts of a given type are indexed based on the combination of fact values which are used in marking of consequences.

Definition 9. The CorrelationSet is a relation on a given pair of types (T, T'), such that CorrelationSet(T, T') is a set of pairs of the form (a, b) where a is a permutation of facts in T and b is a subset of facts in T' such that a and b are eqipotent and there exists an equality constraint of the form $u_1 = v_1 \Box u_2 = v_2 \Box ... u_n = v_n$ where sequence $u_1, u_2, ..., u_n$ is the elements of a arranged in to a fixed order and $v_1, v_2, ..., v_n$ is the sequence b.

Definition 10. The Index Set is a relation on a given type T and set of all types τ which returns subsets of facts in T which must be indexed. $IndexSet(T, \tau)$ returns every subset x of facts of T where there exists a T' such that T' may prepare for T and x is a right-component of $CorrelationSet(T', T)$.

Definition 11. The Implicit Set is a relation on a given type T and set of all types τ which returns a set of facts in T upon which future correlations may depend. $ImplicitSet(T, \tau)$ returns the union of every subset x of facts in T where there exists a T′ such that T may prepare for T′ and x is a left-component in an element of CorrelationSet(T, T′).

> **Input**: Hyper alert stream H, Hyper-alert types τ
> **Output**: All pairs (h′, h) such that h′ prepares for h and both are in H
> foreach h ∈ H (input in ascending time order) do
>     Let T = Type(h);
>     Let i be a index on ImplicitSet(T, τ);
>     if Lookup(i, h) then
>         | Continue with next alert;
>     end
>     foreach index i on IndexSet(T, τ) do
>         Let the set of hyper-alerts R = Lookup(i, h);
>         foreach h′ ∈ R do
>             | Add the pair (h′, h) to Output;
>         end
>     end
>     foreach Type T′ where T may prepare for T′ do
>         foreach Permutation p, index i on CorrelationSet(T, T′) do
>             | Insert(i, Permute(h, p));
>         end
>     end
> end

**Algorithm 1**: The minimal ATG algorithm

## 5 Hypothesising Missing Alerts

Algorithm 1 does not attempt to deal with missing alerts in the input alert stream. What should happen is that for any alert which arrives and is not explained by a prior alert then those alerts are hypothesized with appropriate fact values. This is done recursively until either a hyper-alert type with in-degree zero is found, no facts can be inferred for a hypothesis or until a real alert is found. Only if a real alert is found will the hypothesized sequence be entered in to the correlation graph. If no results are found in the "search for correlations" stage then the current alert is unexplained. Alerts are hypothesized with attributes satisfying each constraint on each incoming edge. Often times only a subset of the fact values may be inferred for a hypothesized alert as not all values have to be referred to in the equality constraints from the attack model.

This leads to a problem when we recurse more than one step. The recursion needs to terminate when a real hyper-alert may prepare for a hypothesized one. There is no guarantee that an index exists for the subset of fact values in the hypothesized alert. Our approach leads us to consider the hypothesizing problem as identical to the correlation problem, except that our hyper-alerts may contain a partial set of attribute values.

A pre-processing step is introduced in which an expanded version of the IndexSet is calculated so that all such partial sets of attribute values are indexed.

Also we intrdouce the relation $HypothesisSet(T, \tau)$ where T is a type and $\tau$ is the set of all hyper-alert types. This relations maps on to a set of 5-tuples with the components (t, i, p, m, o):

1. t is a type which may prepare for T.
2. i is an element of the $IndexSet$ of t.
3. p is a permutation to apply to fact values of the current hyper-alert in order to query the index i of type t.
4. m is the combination of facts which appear in p.
5. o is the combination of facts that were originally required for the current constraint. In other words all facts mentioned on the right hand side of the equality comparisons for this constraint.

The hypothesizing algorithm then is a recursive procedure with two parameters the first of which is a TG vertex $v'$ and the second is a hyper-alert h. The function returns true if a real hyper-alert was correlated or false otherwise. The procedure is that for each element in the $HypothesisSet$ associated with $v'$:

1. Let f be the set of hypothesized fact attributes in h. Continue the loop if the union of f and o is not equal to m. This avoids generating unnecessary hypotheses based on a strict subset of the actually available fact attributes.
2. Let $h'$ be a new hyper-alert. Use p to permute the facts in h and assign them to $h'$.
3. Create a key from $h'$ which combines facts required for index i of t. Query i and if a result is found, correlate the result with $h'$ and continue the loop.
4. Recurse to the vertex for type t passing hyper-alert $h'$. If the recursion returns true then correlate $h'$ with h.

With this procedure hyper-alerts with identical attributes may be created in order to satisfy different paths through the attack graph even though they may eventually lead to the same place. Such alerts add nothing to the intelligibility of the result since one real alert could conceivably account for all such identical hypotheses.

We define two hyper-alerts as strategically indistinguishable provided that they are of the same type, have the same combination of facts assigned with the same values and appear before the hyper-alert they have been hypothesized to explain. Similarly to the implicit correlation step described in the previous section a hypothesized alert database is added to each vertex in the type-graph.

## 6 Empirical Results

To verify the theory the algorithm is implemented in C[3]. Trivial sub-graph elimination is implemented by keeping count of vertex degress in CG as edges are added, only vertices with degree greater than zero are output. This small addition makes output graphs more manageable. The Lincoln Labs 1.0 dataset is used in the experiments for the purposes of generating results comparable to prior works. These data-sets include labeling data which allows for the construction of a perfectly accurate series of alerts. An attack model almost identical to that in [11] is used. The only difference is in fixing an error in the original in which UDP port-scans could be said to discover TCP services and vice versa, which is not the case. All experiments were run on a PC with 1.6GHz Intel Core 2 Duo CPU and 1GB RAM running a contemporary Linux distribution.

Two experiments are proposed: experiment #1 is designed to verify that the algorithm is suitable for application in a real-time correlation setting as intended. Experiment #2 is designed to qualitatively assess the hypothesizing algorithm when a random subset of relevant alerts have been deleted from a perfectly accurate alert stream.

### 6.1 Performance

The aim of this experiment is to test the suitability of our algorithm for real-time correlation. The method is to intersperse a true scenario consisting of 855 alerts within a large number of randomly generated alerts such that there are 1,000,000 alerts in total. No direct comparison with prior work is possible here since comparable algorithms are either not intended for real-time setting, do not perform the hypothesizing step or use a different attack model. Instead, the time taken for the software to perform the work will be recorded and divided by the number of alerts which will give us a correlation-rate. As long as the correlation-rate is higher than the rate at which we expect alerts to be produced by the underlying IDS then the algorithm ought to be suitable for real-time operation. The size of the output graphs is also recorded representing the bulk of the memory utilization of the program.

There are several parameters in this experiment. Firstly we will run the experiment with variations of the algorithm so that we can get an idea of the costs and benefits of each.

1. Algorithm 1. Minimal IAC problem.
   (a) With implicit correlations disabled.
   (b) With implicit correlations enabled.
2. Algorithm 2. Extended IAC problem.
   (a) Without consolidating strategically indistinguishable hypotheses.
   (b) Strategically indistinguishable hypotheses consolidated.

The question arises of how precisely to generate a large number of randomized false positive alerts. The attribute space is 96 bits in total, based on two 32 bit IP addresses, and two 16 bit port numbers. If all attributes are totally randomized the probability of false correlations being generated is exceedingly small. Conversely if we devise a non-random worst-case data-set in which false alarms are crafted specifically to generate correlations then we are venturing in to the area of specific attacks aimed at the correlator itself which is a problem beyond the scope of this paper.

The chosen solution is based on the observation that in a real-world setting the IDS is most often connected to a point in an IP network where it can observe all traffic entering or leaving that network. Therefore while one out of the source and destination addresses of a packet may be any of $2^{32}$ possible IP address values, the other side will be set to one of the addresses on the monitored network which will be a small subset of that address space. Traffic not conforming to these rules is taking place outside of the range of communications systems that the underlying IDS is placed to observe. Similarly, IP services tend to listen on well known ports, typically those under 1024.

Two randomization methods are chosen, one based on a class C IP network and the other on a class B network. These types of networks are defined as having $2^8$ and $2^{16}$ addresses each. The algorithm for generating the data is:

1. Pick a totally random IP address and port number.
2. Pick a random IP address within the allowable range of our network class.
3. Pick a random 10 bit port number.
4. Toss a coin, if heads then the fully random IP is the source address, else it's the destination address.
5. Toss another coin, if heads then the fully random port number is the source address, else vice versa.

Five versions of the random data-set are created for each type of network, making ten data sets in total. Each of the four variations of the algorithm were run on each of the 10 data-sets making 40 runs in total. Each run is repeated three times and the mean CPU time taken as the final result. The variation in run time on the program on the same data-set turned out extremely low so, for the sake of concision, the individual run-timings are not presented here. The 885 real alerts from the LLDOS labeling data are interspersed randomly, but correctly ordered, within each dataset.

Table 1. CPU Times for Class B and Class C Respectively.

| Class | Exp. | Min. (s) | Max. (s) | Mean (s) | Mean Rate (a/s) |
|---|---|---|---|---|---|
| B | 1(a) | 7.47667 | 9.21 | 7.905 | 126,502 |
|   | 1(b) | 5.31 | 6.26333 | 5.675 | 176,221 |
|   | 2(a) | 6.55333 | 6.67667 | 6.599 | 151,541 |
|   | 2(b) | 6.45333 | 6.48 | 6.461 | 154,772 |
| C | 1(a) | 7.0533 | 7.14667 | 7.088 | 141,088 |
|   | 1(b) | 6.79667 | 6.87 | 6.818 | 146,675 |
|   | 2(a) | 11.46 | 11.6033 | 11.52 | 86.380 |
|   | 2(b) | 10.9067 | 19.9233 | 12.43 | 80,440 |

If we look at the final column of table 1 we can observe that the correlation rate is on the order of 100,000 alerts per second. This seems likely to be much faster than an IDS, certainly the majority of deployments in any case.

In table 2 the columns stand for the total number of vertices and edges in the output CG respectively. The number of false alerts seems rather alarming considering only 885 of them are part of our scenario. Although, bear in mind that our noise is distributed over only 20 alert types which are quite highly connected. Further we have opted to restrict alert values to "realistic" ranges. In practice a million alerts do not occur over a few seconds but perhaps days or weeks.

Table 2. Output size for Class B and Class C Respectively.

| Exp. | Hyper-Alerts | Correlations | Hyper-Alerts | Correlations |
|---|---|---|---|---|
| | Class B | | Class C | |
| 1(a) | 194,817 | 157,734 | 346,782 | 888,262 |
| 1(b) | 182,727 | 148,457 | 129,220 | 641,115 |
| 2(a) | 376,786 | 401,974 | 190,986 | 691,809 |
| 2(b) | 299,395 | 302,553 | 190,417 | 691,112 |

## 6.2 Quality of Output

The aim of this experiment is to take the same totally accurate data-set and remove random alerts and test the accuracy of hypothesizing by how accurate the the graphs are as an increasing number of alerts are missed. Unfortunately the number of ways of doing this with a data-set of of 855 alerts, such as ours, is astronomical and our sample sizes would have to be inappropriately large to gain results which can be interpreted with any confidence. From experience the algorithm is extremely robust either when all alerts of one or two types are removed or scores of alerts removed randomly. This intuition leads us on to an alternative experimental setup. There are only four types of alerts in the experimental data set. At least two types are required for there to be correlations and if all alerts are present then the output is ideal. We shall experiment with removing all 2 and 3 combinations of alert types and examining the false correlation rates which are calculated by hand in this case.

These experiments are run with Algorithm 2(b) only. To calculate false alert rates the output graphs are compared against the complete correlation graph which contains 58 hyper-alerts. A false negative is counted for every alert in the complete CG for which no hypothesis exists. Conversely a false positive is counted for every hypothesis which does not correspond to a hyper-alert in the complete CG. For labeling purposes alert types are named A, B, C and D, standing for ping-sweep, sadmind-ping, sadmind-exploit and mstream-zombie respectively.

The results in table 3 are difficult to analyze without taking a closer look at the output graphs produced. For attack sequences which are short in length, missing alerts can have a drastic effect on the false negative correlation rates. False positive hypotheses are a slightly less serious problem and in this case would be entirely eliminated with existing audit-record correlation techniques, as proposed in [15].

Table 3. Hypothesis Accuracy.

| Input Types | False Negatives | False Positives |
|---|---|---|
| ABD | 3 | 12 |
| BCD | 32 | 0 |
| ACD | 26 | 0 |
| ABC | 14 | 0 |
| AC | 37 | 0 |
| BD | 35 | 12 |
| CD | 41 | 0 |
| BC | 44 | 0 |
| AD | 35 | 12 |
| AB | 20 | 0 |

## 7  Conclusions and Future Work

In this paper a real-time correlation algorithm using hyper-alert type graphs was proposed. Our general approach was to reduce the minimal IAC problem to a series of insertions to and removals from a balanced binary tree. We proceeded from there to approach the extended (hypothesizing) problem by re-phrasing the minimal problem such that we can recursively input hyper-alerts with unknown (or wild-card) attributes. It was shown that such algorithms are feasible provided a few conditions are met:

– The number of comparable facts in hyper-alerts is small.
– If hyper-alerts are to be hypothesized then type-graphs should be chosen carefully in order to prevent a exponential explosion in time complexity.

The algorithm was implemented and validated through a series of experiments which showed that a good implementation is suitable for real-time correlation even in cases where the IDS alert rate is alarmingly high. In these cases the size of the output graph becomes the overriding factor in determining the practical utility of the algorithm. It was also confirmed that picking the right aggregation function is invaluable in this respect by allowing many hyper-alerts to be merged in to a single logical unit. However it is not immediately clear how best to design these functions such as to minimize large output graphs to a satisfactory degree.

Although our approach does not require a vulnerability assessment it has been shown that it is possible to make use of such information if it is there [15]. It appears that our algorithm could be modified for similar purposes. The basic approach here would be to incorporate special constraints which depend on external evidence sources. These would be checked before correlating or hypothesizing an alert. However this leads to question of how to determine when to ignore false negative vulnerability assessments if a successful attack of the relevant type has been observed? This may also be a fruitful direction for investigation.